\documentclass[runningheads]{llncs}

\usepackage{formatting}

\usepackage{formattingabbrv}

\usepackage{graphicx}
\usepackage{booktabs}
\usepackage{amsmath}
\graphicspath{{figures/}}
\usepackage[nolist]{acronym}
\usepackage{multirow}
\usepackage{algorithm}
\usepackage{algorithmicx}
\usepackage{algpseudocode}
\usepackage{xspace}
\usepackage{pifont}

\usepackage[accsupp]{axessibility}  

\usepackage{hyperref}

\usepackage{orcidlink}

\newcommand{\ra}[1]{\renewcommand{\arraystretch}{#1}}

\algnewcommand{\algorithmicforeach}{\textbf{for each}}
\algdef{SE}[FOR]{ForEach}{EndForEach}[1]
  {\algorithmicforeach\ #1\ \algorithmicdo}
  {\algorithmicend\ \algorithmicforeach}

\algrenewcommand\algorithmicrequire{\textbf{Input:}}
\algrenewcommand\algorithmicensure{\textbf{Output:}}

\DeclareMathOperator*{\argmax}{arg\,max}

\newcommand*\circled[1]{\tikz[baseline=(char.base)]{
            \node[shape=circle,draw,inner sep=1pt] (char) {#1};}}

\newcommand{\cmark}{\ding{51}}%
\newcommand{\xmark}{\ding{55}}%

\begin{document}

\title{Mitigating Backdoor Attacks using Activation-Guided Model Editing} 

\author{Felix Hsieh\inst{1,2}
\and Huy H. Nguyen\inst{1}
\and AprilPyone MaungMaung\inst{1}
\and Dmitrii Usynin\inst{2,3}
\and Isao Echizen\inst{1,4}
}

\authorrunning{Hsieh et al.}

\institute{National Institute of Informatics, Tokyo, Japan \and
Technical University of Munich, Munich, Germany\and
Imperial College London, London, United Kingdom\and
The University of Tokyo, Tokyo, Japan}

\maketitle

\begin{abstract}

Backdoor attacks compromise the integrity and reliability of machine learning models by embedding a hidden trigger during the training process, which can later be activated to cause unintended misbehavior.
We propose a novel backdoor mitigation approach via machine unlearning to counter such backdoor attacks. The proposed method utilizes model activation of domain-equivalent unseen data to guide the editing of the model's weights. Unlike the previous unlearning-based mitigation methods, ours is computationally inexpensive and achieves state-of-the-art performance while only requiring a handful of unseen samples for unlearning.
In addition, we also point out that unlearning the backdoor may cause the whole targeted class to be unlearned, thus introducing an additional repair step to preserve the model's utility after editing the model.
Experiment results show that the proposed method is effective in unlearning the backdoor on different datasets and trigger patterns.

\keywords{Backdoor Mitigation \and Machine Unlearning \and Model Editing.}
\end{abstract}

\section{Introduction}
\label{sec:intro}

Machine learning models highly depend on the quality and quantity of data available during training. As the demand for more powerful models increases, so does the need for vast data collections and significant computational resources for model training. Except for major corporations, most entities rely on uncurated data, such as publicly available data online and third-party services that run learning protocols. The loss of control of the training opens up an attack vector for a malicious actor to use backdoor attacks to poison the training data~\cite{li2022backdoor}.
 
Gu et al.~\cite{gu2019badnets} proposed BadNets, the first backdoor attack. BadNets overlays a small subset of training samples with a square of fixed size and position, and it changes the labels to a target class, thus poisoning the samples. During training, the victim model learns to associate the trigger pattern with the target class, creating a hidden backdoor reactive to the trigger. During inference, the model behaves as usual on clean data. Still, when a malicious actor forwards a sample with a specific trigger, the backdoor in the neural network is activated, leading to model misbehavior, such as misclassification. A survey from Microsoft stated that data poisoning is one of the top attacks on machine learning systems~\cite{kumar2020adversarial}.

The security risk backdoor attacks impose creates the need for contrary defense methods. In this work, we focus on one type of defense where one mitigates the influence of a backdoor attack on an adversarial-modified model. 
Retraining the model from scratch with clean training data is the most straightforward approach for obtaining an adversarial-free model. Retraining is computationally expensive and requires access to clean training data. Filtering out poisoned samples in a training dataset is often unfeasible because the dataset is too large. 
Backdoor mitigation with machine unlearning has emerged as a promising approach to overcoming the limitations of retraining and efficiently removes a backdoor in a poisoned model. 
Many methods omit the need for original training data.

\begin{figure}[tb]
\centering
\includegraphics[width=0.85\linewidth]{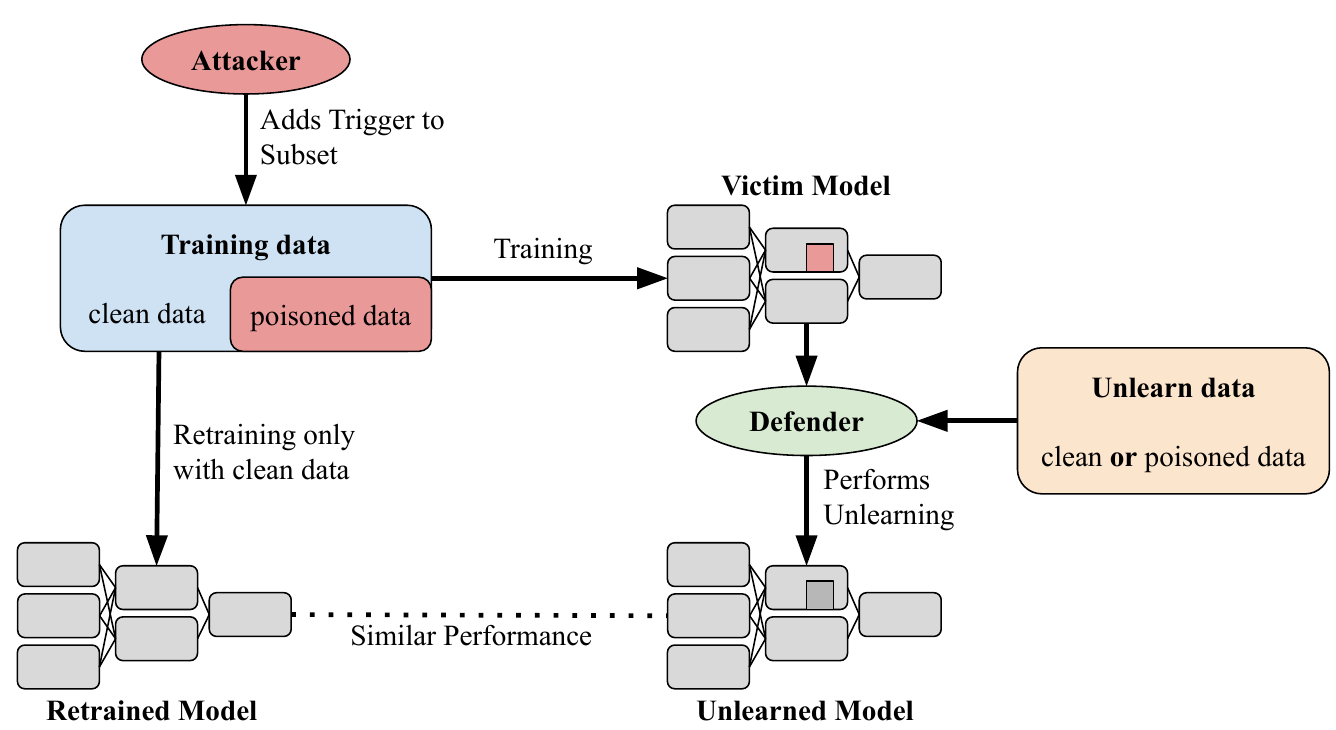}
\caption{Summary of backdoor unlearning setting.\label{fig:setting}}
\end{figure}

This work focuses on a realistic scenario where the defender cannot access the dataset used to train the victim model. Figure~\ref{fig:setting} shows an overview of our unlearning setting. Unlearning aims to obtain a model that performs similarly to a model retrained on a clean subset of training data for clean and poisoned data. For unlearning the backdoor, we have access only to a limited domain-equivalent unseen dataset, the length of which is an order of magnitude smaller than the original training dataset. In practice, collecting a fitting dataset with annotations is expensive and time-consuming. Methods that are effective with an even smaller data count available are of particular interest for this scenario because they allow secure hand-picking of clean data for unlearning. 

We propose a novel backdoor-unlearning approach that uses information from an extracted activation to guide the editing of model weights. This process aims to mitigate the influence of backdoor samples in the training dataset.
Editing the weights is beneficial because it allows us to selectively target and repair the parts compromised by the attack.
In addition to directly editing the weights, unlearning benefits from optionally allowing the parameters of the \ac{bn} layer to be changed during activation extraction. 

The contributions of this work are as follows:
\begin{itemize}
	\item We propose a novel model-editing method for unlearning samples with backdoor triggers by utilizing the activation of clean or poisoned samples extracted for a backdoored model. The proposed method is time- and sample-efficient.
    \item We point out that the proposed unlearning might unlearn the targeted class, thus introducing an optional repair process to preserve utility while forgetting only the backdoor trigger.
	\item We conduct experiments under two scenarios (with or without knowledge of the backdoor trigger) with three state-of-the-art backdoor attacks on different models and datasets. We present the results with an analysis.
\end{itemize}
In the experiments, the proposed method can consistently outperform other baseline methods.


\section{Related Work}
This section briefly reviews backdoor attacks, backdoor defenses, and machine unlearning.

\subsection{Backdoor Attacks}
Backdoor attacks involve preemptively poisoning a subset of training data with a specific backdoor trigger pattern and a target label. During training, a neural network learns that images with a specified trigger correspond to a target class, thus introducing an additional adversarial task. During inference, the network works as usual on benign data. A malicious actor can activate the backdoor to manipulate model response, causing misbehavior, such as misclassifications.

\begin{figure}[tb]
\centering
\subfloat[White]{\includegraphics[width=0.24\linewidth]{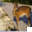}\label{fig:white}}
\hfil
\subfloat[Mean]{\includegraphics[width=0.24\linewidth]{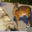}\label{fig:mean}}
\hfil
\subfloat[Apple]{\includegraphics[width=0.24\linewidth]{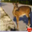}\label{fig:apple}}
\hfil
\subfloat[TEST1]{\includegraphics[width=0.24\linewidth]{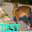}\label{fig:test}}\\
\subfloat[TEST2]{\includegraphics[width=0.24\linewidth]{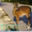}\label{fig:test2}}
\hfil
\subfloat[Gaussian Noise]{\includegraphics[width=0.24\linewidth]{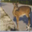}\label{fig:noise}}
\hfil
\subfloat[Invisible]{\includegraphics[width=0.24\linewidth]{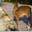}\label{fig:invisible}}
\hfil
\subfloat[Narcissus]{\includegraphics[width=0.24\linewidth]{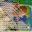}\label{fig:narcissus}}
\caption{Examples of eight backdoor triggers on CIFAR10. Images (a)--(f) are poisoned by BadNets~\cite{gu2019badnets} with different patches, (g) is with Steganography~\cite{li2020invisible}, and (h) is with Narcissus~\cite{zeng2023narcissus}.\label{fig:trigger}}
\end{figure}

There exist various types of backdoor attacks~\cite{li2022backdoor}.
BadNets~\cite{gu2019badnets}, the first backdoor attack, uses noticeable square patches as triggers. 
In contrast to visible triggers, for invisible triggers, poisoned images are indistinguishable from clean ones, as in~\cite{li2020invisible,li2021invisible}. 
Backdoor attacks with optimized triggers~\cite{Liu2018TrojaningAO,zeng2023narcissus} are designed to be more effective and thus usually require fewer poisoned training samples.
Moreover, a shared semantic part of the images can be used as a trigger~\cite{bagdasaryan2020backdoor,lin2020composite} without manipulating the images and only changing the labels.
In addition, instead of using a single trigger pattern, certain methods allow for varying sample-specific triggers~\cite{nguyen2020input}.
Although the targeted label is usually for a single class, there are all-to-all attacks~\cite{gu2019badnets2} that use different target labels.
In this work, we use visible, invisible, and optimized triggers for our experiments, and examples of such triggers are shown in Figure~\ref{fig:trigger}.

\subsection{Backdoor Defenses}
With the development of backdoor attacks, researchers have proposed various backdoor defenses as countermeasure~\cite{li2022backdoor}. Most defense methods should instead be considered mitigation methods because, in most cases, they cannot entirely erase the influence of the attacks. One such mitigation is data pre-processing prior to model inference, which aims to perturb the trigger to not activate the backdoor~\cite{liu2017neural,doan2020februus}. Another form of defense, typically used to aid other defense methods, is trigger synthesis~\cite{wang2019neural}. With reverse engineering, trigger synthesis approximates the trigger, which can be used for trigger-guided defense or to retrieve the target class. Model diagnosis~\cite{kolouri2020universal,xu2020detecting} is another type of defense to detect a backdoor and prevent model deployment. Moreover, poison suppression defenses modify the training process to be robust against backdoor creation~\cite{du2019robust,hong2020effectiveness}. Other methods utilize sample filtering to detect trigger images and remove them from the training set or decline them during model inference~\cite{gao2020strip,tran2018spectral,chen2018detecting}. Some defense methods aim to remove the backdoor from an infected model by directly modifying the model ~\cite{liu2017neural,liu2018finepruning}. In this work, we propose a method that directly edits model weights to erase the backdoor in a poisoned model. The editing uses the activation of clean or poisoned samples as a guide.

\subsection{Machine Unlearning}
The influence of specific training samples on a model can be mitigated with machine unlearning. This influence can be entirely removed by retraining the model from scratch without the data we want to forget. One limitation of this approach is the requirement for training data, which can be inaccessible or too big to filter out the data we want to forget. Another issue is the high computational and time resource expense associated with retraining~\cite{xu2023machine}.

Different machine unlearning methods try to evade some of those limitations~\cite{xu2023machine}. One approach is data obfuscation~\cite{graves2020amnesiac,tarun2023fast}, where the model is fine-tuned with additional obfuscated data that disturbs the functionality of the data we want to forget. Certain approaches require design choices prior to training, like multi-model-aggregation~\cite{bourtoule2020machine,gupta2021adaptive} or a transformation layer inserted between data and model~\cite{cao2015towards}. For specific model manipulation methods, model weights can be shifted by an update value~\cite{guo2019certified,golatkar2020eternal}, replaced by new values~\cite{schelter2021hedgecut,wu2020deltagrad}, or pruned~\cite{wang2022federated,baumhauer2022machine} and usually repaired with a subsequent fine-tuning step. The scope of the information targeted for unlearning can range from whole classes~\cite{tarun2023fast,shibata2021learning} to individual samples~\cite{graves2020amnesiac}. This work focuses on backdoor attacks and aims to unlearn the features of a backdoor trigger pattern learned by the victim model.

\section{Methodology}

We consider an adversarial-modified (backdoored) image classifier $f_\theta$ parameterized by $\theta$, which is trained with a dataset $D$ that is comprised of clean data and backdoored data ($D = {D_C \cup D_B}$). Samples in $D_B$ contain the backdoor trigger $\delta$ and have the target label $y_t$.
$D_B$ is usually a small fraction of a clean training set $D_T$ with a budget $\rho$ such that $|D_B| \leq \rho|D_T|$.
$f_\theta$ takes an input image $x \in \mathcal{X}$ and $f_\theta(x)_i$ represents the probability that $x$ corresponds to label $i \in \mathcal{Y}$. $\mathcal{X}$ is the input space, and $\mathcal{Y}$ is the label space. The predicted label $\hat{y}$ is obtained by using the arg max operation ($\argmax_i f_\theta(x)_i$). 
Since $f_\theta$ is backdoored, $f_\theta$ works as normal on a clean input $x_c$ (\ie predicting $\hat{y}$) and predicts $y_t$ for input $x_b$ embedded with the backdoor trigger $\delta$.
We aim to unlearn $D_B$ that $f_\theta$ does not predict $y_t$ when given $x_b$.
Here, we slightly abuse the notation and imply that $f_\theta$ is a deep neural network with multiple layers.
Specifically, we consider a neural network with multiple blocks of convolutional layers with or without \ac{bn}.

Given $f_\theta$ without having access to the training dataset $D$, we propose an activation-guided model editing approach to unlearn $D_B$ under two assumptions: (1) we have \ac{bdk}, and (2) we do not have it ($\neg$\ac{bdk}). 
For both assumptions, we split the total weights of $f_\theta$ into two halves and add those layers corresponding to the weights of the second half to a layer list $L$, for which we want to edit the weights. We target those later layers because they have the highest proximity to the classification output. We do not want to edit the early layers associated with general low-level feature extraction~\cite{GoodBengCour16}. The authors of other backdoor mitigation approaches also suggest that focusing the unlearning on the later layers improves performance~\cite{liu2018finepruning, wang2019neural}.
First, we prepare an unlearning dataset $D_U$ with the same distribution as the training dataset $D$. However, $D_U$ is not used in training $f_\theta$. Our empirical experiments suggest that $D_U$ can be as small as four samples.

\begin{figure}[tb]
\centering
\includegraphics[width=0.85\linewidth]{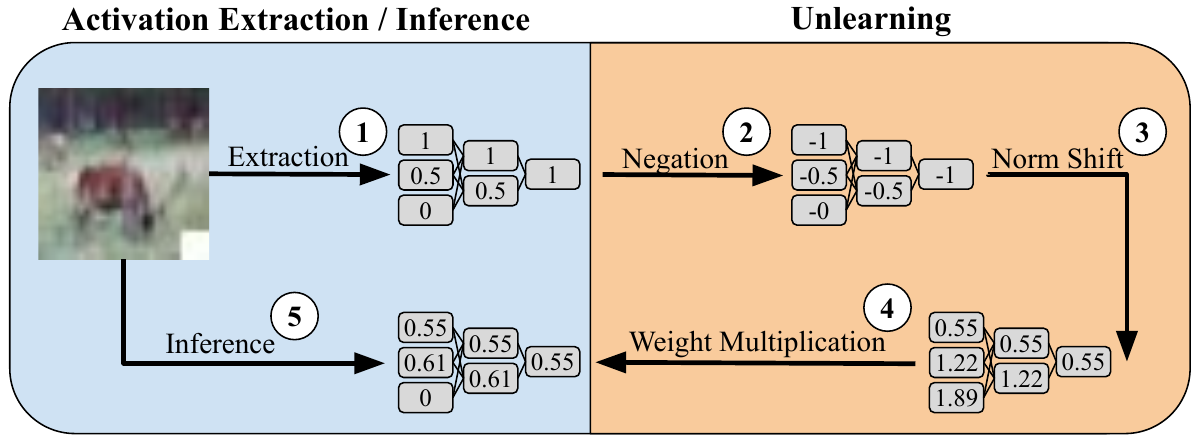}
\caption{Overview of proposed activation-guided model editing approach.\label{fig:method_example}}
\end{figure}

\subsection{Assumption 1 - BDK}
As we assume we have backdoor information in this scenario, we poison $D_U$ with the known backdoor trigger $\delta$.
In addition, for models with \ac{bn}, we freeze \ac{ma} parameters during activation extraction because this was experimentally proven to be more effective for unlearning. 
Figure~\ref{fig:method_example} shows an overview of the proposed model editing process, 
which works as follows.

\vspace{2mm}\noindent{\bf \circled{1}}
Extract model activation $A$ for the whole dataset $D_U$. Therefore, iterate batch-wise over data $X$ in $D_U$ infected with trigger $\delta$ and forward it through model $f_\theta$ given $f_\theta(X + \delta)$. Capture and average the activations across all batches for each layer $l$ in layer list $L$. 

\vspace{2mm}\noindent{\bf \circled{2}}
Next, negate each value in the activation $A$ as
\begin{equation}\label{eq:neg}
    A = -A.
\end{equation}

Then, iterate over $l$ in list $L$ of the layers targeted for editing. For each layer $l$, check if the later multiplication of weights $\theta_l$ and activation $A_l$ would result in a matrix-multiplication-error caused by shape mismatch. If this is the case, use adaptive average pooling to adjust the shape of $A_l$ to match the shape of $\theta_l$. 

\vspace{2mm}\noindent{\bf \circled{3}}
Then, compute layer-wise mean and standard deviation statistics as
\begin{equation}
    \mu_l =  \frac{1}{m} \sum_{i=1}^{m} a_i, \quad \sigma^2_l = \frac{1}{m} \sum_{i=1}^{m} (a_i - \mu_l)^2,
\end{equation}
where $a_i \in A_l$.
Then, use the calculated statistics to normalize activation $A_l$ for each layer $l$ as
\begin{equation}
    A_l = \frac{A_l - \mu_l}{\sqrt{\sigma^2_l + \epsilon}},
\end{equation}
where $\epsilon$ is a small value to avoid division by zero.
Next, rescale $A_l$ with scale and shift hyperparameters $\gamma$ and $\lambda$ as
\begin{equation}\label{eq:rescale}
    A_l = \gamma \cdot A_l + \lambda.
\end{equation}

\vspace{2mm}\noindent{\bf \circled{4}}
With the calculated activation factor $A_l$, edit the model weights $\theta_l$ as
\begin{equation}
    \theta_l = \theta_l \cdot A_l.
\end{equation}
Modifying $\theta$ for all layers in $L$ completes the unlearning process. We obtain an unlearned model with a mitigated backdoor.

\vspace{2mm}\noindent{\bf \circled{5}}
Finally, evaluate the classification performance on a separate verification test set $D_V$ to evaluate the unlearning process.
\vspace{2mm}

In Equation~\ref{eq:rescale}, we scale and shift the activation with values 0.5 and 1.0 for $\gamma$ and $\lambda$ hyperparameters, which changes the activation mean to 1.0. With this, we aim to preserve the utility of the model during unlearning and mitigate the internal covariate shift, especially after significant distribution changes by Equation~\ref{eq:neg}. Choosing those values should lead to minimal change in the inherent mean of the weights when multiplying with the shifted activation. We experimentally confirmed the performance with those values in the supplementary material.

After model editing, we observe that the model tends to unlearn the whole class instead of unlearning backdoored samples only. To address this issue, we introduce an optional repair phase to restore some model utility if we have unlearned more than intended. Specifically, we fine-tune the model on the samples in $D_U$ and on the same samples poisoned with $\delta$ for one epoch each, using the correct ground truth as the target label in both cases.

\subsection{Assumption 2 - $\neg$BDK}
In this scenario, we do not have any information about the backdoor trigger or algorithm ($\neg$\ac{bdk}).
Therefore, we cannot poison the unlearning dataset $D_U$ and use the clean $D_U$ as it is in the unlearning process.
For model editing with $\neg$\ac{bdk}, we perform the process in Figure~\ref{fig:method_example} with two modifications: (1) backdoor trigger $\delta$ is zero as we do not have any information about it, and (2) we update \ac{ma} parameters during activation extraction to aid unlearning of dataset $D_U$ for models with \ac{bn} layers, unlike the unlearning process with \ac{bdk}.
We perform the optional repairing by fine-tuning on the clean $D_U$ set.

\begin{table}[htbp]
\caption{Evaluation of proposed unlearning on five different datasets. Additional repairing was performed with learning rate value of $1\mathrm{e}{-2}$. We compare base state after unlearning with one after repairing. We used Top-5 accuracy for CIFAR100 and TinyImageNet. Best results are highlighted in bold.}
\centering
\ra{1.1}
\resizebox{\linewidth}{!}{%
\begin{tabular}{c@{\hskip 0.2in}c@{\hskip 0.2in}c@{\hskip 0.2in}c@{\hskip 0.2in}c@{\hskip 0.2in}c}\toprule
        \textbf{Dataset}&\multirow{2}{*}{BDK}&\multirow{2}{*}{State} & \multicolumn{3}{c}{Metric}\\
        \cmidrule(lr){4-6} 
        (ASR/ACC)&&& ASR ($\downarrow$) & ACC ($\uparrow$) & CTCA ($\uparrow$)\\
        \midrule
        &\multirow{2}{*}{\cmark} & Base & 0.0$\pm$0.0 & 86.04$\pm$1.2 & 0.0$\pm$0.0\\ 
        MNIST&& +repair & 0.41$\pm$0.16 & \textbf{97.77$\pm$0.37} & \textbf{96.09$\pm$2.51}\\
        \cmidrule(lr){2-6} 
        (99.99/98.98) &\multirow{2}{*}{\xmark} & Base & 66.81$\pm$5.87 & 77.37$\pm$2.84 & 59.19$\pm$11.56\\ 
        && +repair & 67.3$\pm$12.4 & \textbf{97.31$\pm$1.27} & \textbf{98.13$\pm$0.31}\\
        \midrule
        &\multirow{2}{*}{\cmark}& Base & 0.0$\pm$0.0 & \textbf{65.49$\pm$0.39} & 0.0$\pm$0.0\\
        CIFAR10 && +repair & 9.79$\pm$0.88 & 59.87$\pm$3.04 & \textbf{42.07$\pm$7.72}\\
        \cmidrule(lr){2-6} 
        (94.53/70.5) &\multirow{2}{*}{\xmark}& Base &\textbf{0.3$\pm$0.21} & 59.18$\pm$2.63 & 0.14$\pm$0.19\\
        && +repair & 23.09$\pm$8.07 & 61.83$\pm$1.33 & \textbf{55.02$\pm$4.41}\\
        \midrule
        &\multirow{2}{*}{\cmark}& Base & 0.92$\pm$0.33 & 50.68$\pm$1.36 & 3.91$\pm$2.8\\
        CIFAR100&& +repair & 4.13$\pm$1.61 & \textbf{60.42$\pm$0.96} & \textbf{41.27$\pm$10.27}\\
        \cmidrule(lr){2-6} 
        (93.84/63.06) &\multirow{2}{*}{\xmark}& Base  & \textbf{0.0$\pm$0.0} & 37.88$\pm$2.19 & 0.0$\pm$0.0\\
        && +repair & 20.3$\pm$14.16 & \textbf{60.83$\pm$0.81} & \textbf{58.78$\pm$12.86}\\
        \midrule
        &\multirow{2}{*}{\cmark}& Base & 0.0$\pm$0.0 & \textbf{53.08$\pm$0.49} & 0.0$\pm$0.0 \\
        CINIC10&& +repair & 0.0$\pm$0.0 & 10.0$\pm$0.0 & 0.0$\pm$0.0 \\
        \cmidrule(lr){2-6} 
        (96.28/57.35) &\multirow{2}{*}{\xmark}& Base &  0.01$\pm$0.01 & \textbf{41.89$\pm$0.83} & 0.08$\pm$0.1\\
        && +repair & 0.0$\pm$0.0 & 10.0$\pm$0.0 & 0.0$\pm$0.0\\
        \midrule
        &\multirow{2}{*}{\cmark} & Base & 0.48$\pm$0.32 & 34.21$\pm$1.0 & 1.59$\pm$2.24\\
        TinyImageNet&& +repair & 0.86$\pm$0.33 & \textbf{44.33$\pm$0.16} & 4.14$\pm$3.75\\
        \cmidrule(lr){2-6} 
        (95.25/45.21)&\multirow{2}{*}{\xmark} & Base &0.0$\pm$0.0 & 20.26$\pm$1.87 & 0.0$\pm$0.0\\
        && +repair & 0.72$\pm$0.6 & \textbf{44.03$\pm$0.21} & \textbf{7.72$\pm$5.71}\\
        \bottomrule
    \end{tabular}
}
\label{tab:datasets}
\end{table}

\begin{table}[ht]
\caption{Evaluation of proposed unlearning on different models. Additional repairing was performed with learning rate value of $1\mathrm{e}{-2}$. We compare base state after unlearning with one after repairing. Best results are highlighted in bold.}
\centering
\ra{1.1}
\resizebox{\linewidth}{!}{%
\begin{tabular}{c@{\hskip 0.2in}c@{\hskip 0.2in}c@{\hskip 0.2in}c@{\hskip 0.2in}c@{\hskip 0.2in}c}\toprule
        \textbf{Model}&\multirow{2}{*}{BDK}&\multirow{2}{*}{State} & \multicolumn{3}{c}{Metric}\\
        \cmidrule(lr){4-6} 
        (ASR/ACC)&&& ASR ($\downarrow$) & ACC ($\uparrow$) & CTCA ($\uparrow$)\\
        \midrule
        &\multirow{2}{*}{\cmark} & Base & \textbf{0.0$\pm$0.0} & \textbf{65.49$\pm$0.39} & 0.0$\pm$0.0\\
        ResNet18&& +repair & 9.79$\pm$0.88 & 59.87$\pm$3.04 & \textbf{42.07$\pm$7.72}\\
        \cmidrule(lr){2-6} 
        (94.53/70.5) &\multirow{2}{*}{\xmark} & Base & \textbf{0.3$\pm$0.21} & 59.18$\pm$2.63 & 0.14$\pm$0.19\\
        && +repair & 23.09$\pm$8.07 & 61.83$\pm$1.33 & \textbf{55.02$\pm$4.41}\\
        \midrule
        &\multirow{2}{*}{\cmark}& Base & 0.0$\pm$0.0 & 71.23$\pm$0.38 & 0.0$\pm$0.0 \\
        VGG16&& +repair & 4.36$\pm$1.15 & 70.15$\pm$2.71 & \textbf{65.5$\pm$4.69}\\
        \cmidrule(lr){2-6} 
        (96.25/78.39) &\multirow{2}{*}{\xmark}& Base & \textbf{0.0$\pm$0.0} & 70.13$\pm$1.74 & 0.0$\pm$0.0\\
        && +repair & 92.17$\pm$1.17 & 72.23$\pm$0.32 & \textbf{57.83$\pm$7.3}\\
        \midrule
        &\multirow{2}{*}{\cmark}& Base & 0.0$\pm$0.0 & \textbf{47.49$\pm$2.79} & 0.0$\pm$0.0 \\
        EfficientNetV2-S&& +repair & 6.82$\pm$2.96 & 31.51$\pm$5.45 & \textbf{7.94$\pm$3.66} \\
        \cmidrule(lr){2-6} 
        (89.74/48.83) &\multirow{2}{*}{\xmark}& Base &  0.0$\pm$0.0 & 11.07$\pm$1.5 & 0.0$\pm$0.0\\
        && +repair & 1.71$\pm$2.02 & \textbf{31.41$\pm$6.19} & \textbf{4.6$\pm$4.35}\\
        \midrule
        MobileNetV3 &\multirow{2}{*}{\cmark}& Base & \textbf{0.0$\pm$0.0} & \textbf{55.8$\pm$1.84} & 0.0$\pm$0.0 \\
        (small) && +repair & 26.28$\pm$7.53 & 50.63$\pm$3.63 & \textbf{49.18$\pm$11.63}\\
        \cmidrule(lr){2-6} 
        (95.67/62.3) &\multirow{2}{*}{\xmark}& Base & \textbf{6.09$\pm$4.54} & 54.35$\pm$2.93 & 4.19$\pm$2.66\\
        && +repair & 84.69$\pm$6.09 & 51.05$\pm$3.68 & \textbf{46.51$\pm$15.49}\\
        \bottomrule
    \end{tabular}
    }
\label{tab:models}
\end{table}

\begin{table}[ht]
\caption{Score ($\uparrow$) comparison of proposed method with state-of-the-art unlearning methods on different infected backdoors. For backdoor verification, $alpha$ value is multiplied by three up to maximum of 100\%. Best results for each trigger are highlighted in bold.}
\centering
\ra{1.1}
\resizebox{\linewidth}{!}{%
\begin{tabular}{lcccccc}\toprule
        \multirow{2}{*}{\textbf{Infected Trigger}}&poisoned&&& \multicolumn{3}{c}{BDK}\\
        \cmidrule(lr){5-7}
        &(ASR/ACC)&$\rho$& $alpha$&actFT~\cite{qiao2019defending} & BaEraser~\cite{liu2022backdoor} & Ours\\
        \midrule
        (a) White~\cite{gu2019badnets}&(94.53/70.5)  &5\%&1.0& \textbf{94.62$\pm$2.61}&59.5$\pm$4.62&93.02$\pm$0.66\\
        (b) Mean &(88.49/68.71) & 10\%& 1.0& 61.81$\pm$6.09& 70.85$\pm$7.33& \textbf{95.23$\pm$1.79}\\
        (c) Apple &(99.23/69.63) & 5\%& 1.0& 85.21$\pm$11.5& 62.87$\pm$11.29& \textbf{92.36$\pm$2.11}\\
        (d) TEST1 &(99.75/70.73) & 5\%& 1.0& 90.69$\pm$2.43& 36.99$\pm$8.55& \textbf{92.19$\pm$0.87}\\
        (e) TEST2 &(99.99/69.66) & 5\%& 0.15& 88.3$\pm$1.19& 48.57$\pm$19.28& \textbf{91.31$\pm$1.41}\\
        (f) Gaussian Noise &(85.77/68.19) & 5\%& 0.25& 52.97$\pm$2.19& 89.6$\pm$1.56& \textbf{93.57$\pm$0.35}\\
        \midrule
        (g) Invisible~\cite{li2020invisible} &(97.97/61.4) & 50\%& -& 84.99$\pm$3.83& 40.51$\pm$12.39& \textbf{92.24$\pm$2.38}\\
        \midrule
        (h) Narcissus~\cite{zeng2023narcissus} &(99.32/63.72) & 5\%& 0.2&  4.09$\pm$1.61& 51.35$\pm$14.04& \textbf{95.45$\pm$3.1} \\
        \bottomrule
    \end{tabular}
    }
\label{tab:trigger1}
\end{table}

\begin{table}[ht]
\caption{Score ($\uparrow$) comparison of proposed method with state-of-the-art unlearning methods on different infected backdoors. For backdoor verification, $alpha$ value is multiplied by three up to maximum of 100\%. Best results for each trigger are highlighted in bold.}
\centering
\ra{1.1}
\resizebox{\linewidth}{!}{%
\begin{tabular}{lcccccc}\toprule
        \multirow{2}{*}{\textbf{Infected Trigger}}&poisoned&&& \multicolumn{3}{c}{$\neg$BDK}\\
        \cmidrule(lr){5-7}
        &(ASR/ACC)&$\rho$& $alpha$& basicFT & NAD~\cite{li2021neural} & Ours\\
        \midrule
        (a) White~\cite{gu2019badnets}&(94.53/70.5)  &5\%&1.0& 61.9$\pm$3.63& 65.33$\pm$3.04& \textbf{83.73$\pm$3.29}\\
        (b) Mean &(88.49/68.71) & 10\%& 1.0& 50.01$\pm$25.7& 71.79$\pm$2.52& \textbf{89.95$\pm$1.73}\\
        (c) Apple &(99.23/69.63) & 5\%& 1.0& 69.1$\pm$2.7& 70.86$\pm$0.15& \textbf{80.41$\pm$3.78}\\
        (d) TEST1 &(99.75/70.73) & 5\%& 1.0& 31.44$\pm$23.82& 62.26$\pm$3.38& \textbf{83.4$\pm$3.89}\\
        (e) TEST2 &(99.99/69.66) & 5\%& 0.15& 48.56$\pm$14.03& 68.17$\pm$4.35& \textbf{82.47$\pm$0.65}\\
        (f) Gaussian Noise &(85.77/68.19) & 5\%& 0.25& 41.92$\pm$22.62& 31.19$\pm$13.93& \textbf{82.81$\pm$0.77}\\
        \midrule
        (g) Invisible~\cite{li2020invisible} &(97.97/61.4) & 50\%& -& 58.66$\pm$14.52& 73.7$\pm$1.1& \textbf{87.26$\pm$0.58}\\
        \midrule
        (h) Narcissus~\cite{zeng2023narcissus} &(99.32/63.72) & 5\%& 0.2&38.52$\pm$16.51& \textbf{75.75$\pm$1.38}& 17.73$\pm$11.77\\
        \bottomrule
    \end{tabular}
    }
\label{tab:trigger2}
\end{table}

\begin{table}[htbp]
\caption{Performance of proposed unlearning when using different numbers of samples for unlearning. Results represent the model state after unlearning without repairing. Best results are highlighted in bold.}
\centering
\ra{1.1}
\resizebox{\linewidth}{!}{%
\begin{tabular}{c@{\hskip 0.1in}c@{\hskip 0.1in}c@{\hskip 0.1in}c@{\hskip 0.1in}c@{\hskip 0.1in}c@{\hskip 0.1in}c}
\toprule
\textbf{Number of} &\multicolumn{3}{c}{BDK}&\multicolumn{3}{c}{$\neg$BDK}\\
\cmidrule(lr){2-4}\cmidrule(lr){5-7}
\textbf{samples} & ASR ($\downarrow$) & ACC ($\uparrow$) & CTCA ($\uparrow$)& ASR ($\downarrow$) & ACC ($\uparrow$) & CTCA ($\uparrow$)\\
\midrule
2 & 0.0$\pm$0.0 & 63.92$\pm$1.91& 0.0$\pm$0.0& 24.6$\pm$17.84 & 61.1$\pm$1.94& \textbf{6.48$\pm$9.02}\\
4 & 0.0$\pm$0.0 & 65.18$\pm$0.91& 0.0$\pm$0.0& \textbf{0.92$\pm$1.15} & 60.77$\pm$2.25& 0.0$\pm$0.0\\
8 & 0.0$\pm$0.0 & 65.14$\pm$0.65& 0.0$\pm$0.0& 5.76$\pm$6.26 & 60.19$\pm$2.96& 0.0$\pm$0.0\\
16 & 0.0$\pm$0.0 & 65.17$\pm$0.68& 0.0$\pm$0.0 & 4.82$\pm$6.75 & 60.77$\pm$2.91& 0.0$\pm$0.0\\
32 & 0.0$\pm$0.0 & 65.47$\pm$0.44& 0.0$\pm$0.0& 9.26$\pm$12.93 & \textbf{63.77$\pm$0.58}& 0.07$\pm$0.1\\
64 & 0.0$\pm$0.0 & 64.7$\pm$1.45& 0.0$\pm$0.0& 7.63$\pm$8.32 & 61.9$\pm$2.22& 1.94$\pm$2.74\\
128 & 0.0$\pm$0.0 & 65.53$\pm$0.45& 0.0$\pm$0.0& 4.38$\pm$5.58 & 61.67$\pm$1.22& 0.07$\pm$0.1\\
256 & 0.0$\pm$0.0 & 65.51$\pm$0.41& 0.0$\pm$0.0& \textbf{0.97$\pm$1.02} & 58.86$\pm$2.42& 0.2$\pm$0.16\\
512 & 0.0$\pm$0.0 & 65.49$\pm$0.39& 0.0$\pm$0.0 & \textbf{0.3$\pm$0.21} & 59.18$\pm$2.63& 0.14$\pm$0.19\\
5000 & 0.0$\pm$0.0 & 64.57$\pm$1.53& 0.0$\pm$0.0& 10.48$\pm$11.24 & 61.41$\pm$0.82& \textbf{6.75$\pm$9.54}\\
\bottomrule
\end{tabular}
}
\label{tab:samplecount}
\end{table}

\begin{table}[ht]
\caption{Efficiency of proposed unlearning compared with state-of-the-art methods. Experimented with 50\%(5000), 5\%(500), 0.5\%(50), and 0.05\%(5) of unseen CIFAR10 data for unlearning. Table displays only sample runs with highest and second-highest scores. Full table is displayed in supplementary material. Best results are highlighted in bold.}
\centering
\ra{1.1}
\resizebox{\linewidth}{!}{%
\begin{tabular}{lcccccc}\toprule
\multirow{2}{*}{\textbf{Method}} &Number of& \multicolumn{5}{c}{Metric} \\
\cmidrule(lr){3-7}
&Samples& Score ($\uparrow$) & ASR ($\downarrow$) & ACC ($\uparrow$)& CTCA ($\uparrow$) & Time ($\downarrow$)\\
\midrule
actFT~\cite{qiao2019defending}&5000& \textbf{92.95$\pm$4.72} & 5.28$\pm$2.48 & \textbf{69.64$\pm$0.92} & \textbf{58.1$\pm$9.1} & 3.96$\pm$0.34\\
&500& 7.0$\pm$2.09 & 87.17$\pm$2.86 & 69.41$\pm$1.02 & 61.47$\pm$7.99 & 2.81$\pm$0.04\\
\cmidrule(lr){2-7}
BaEraser~\cite{liu2022backdoor}& 500 & 80.69$\pm$1.93 & 3.81$\pm$2.41 & 59.45$\pm$0.83 & 32.98$\pm$1.1 & 138.88$\pm$2.87\\
&5000& 57.47$\pm$15.65 & 2.0$\pm$2.79 & 41.81$\pm$12.3 & 14.79$\pm$20.77 & 623.3$\pm$111.46\\
\cmidrule(lr){2-7}
Ours(BDK)& 50 & \textbf{92.4$\pm$1.39} & \textbf{0.0$\pm$0.0} & 65.29$\pm$0.58 & 0.0$\pm$0.0 & \textbf{0.38$\pm$0.01} \\
&5& 92.51$\pm$1.34 & 0.0$\pm$0.0 & 65.37$\pm$0.56 & 0.0$\pm$0.0 & 0.38$\pm$0.02\\
\cmidrule(lr){1-7}
basicFT&5000& 69.49$\pm$1.11 & 6.0$\pm$1.07 & 52.57$\pm$0.91 & \textbf{33.86$\pm$4.07} & 71.56$\pm$1.47\\
&5& 14.17$\pm$0.11 & 0.0$\pm$0.0 & 10.01$\pm$0.0 & 0.0$\pm$0.0 & 71.49$\pm$0.03\\
\cmidrule(lr){2-7}
NAD~\cite{li2021neural}& 5000& 71.23$\pm$3.2 & 4.54$\pm$1.48 & 52.95$\pm$1.41 & 32.05$\pm$1.3 & 114.9$\pm$1.45\\
&500& 42.18$\pm$5.42 & 6.68$\pm$2.06 & 32.09$\pm$3.97 & 23.18$\pm$9.4 & 117.69$\pm$1.9\\
\cmidrule(lr){2-7}
Ours($\neg$BDK)& 50 & \textbf{89.9$\pm$2.02} & \textbf{0.0$\pm$0.0} & \textbf{63.52$\pm$1.24} & 0.0$\pm$0.0 & \textbf{0.36$\pm$0.02} \\
&5& 89.5$\pm$1.72 & 0.0$\pm$0.0 & 63.23$\pm$0.86 & 0.0$\pm$0.0 & 0.4$\pm$0.01\\
\cmidrule(lr){1-7}
Retraining&47500 & -& 4.03$\pm$0.99 & 70.27$\pm$ 0.66 & 	60.55$\pm$4.3 & 423.56$\pm$73.96\\
\bottomrule
\end{tabular}
}
\label{tab:time}
\end{table}

\section{Experiments}

In this section, we perform various experiments to show the effectiveness of our method in different settings and compare it with other existing backdoor unlearning methods. 
We conducted all experiments three times, and the averaged results are summarized as follows.

\subsection{Setup\label{sec:setup}}

\noindent{\bf Datasets.}
We explored our method on MNIST~\cite{lecun1998gradient}, CIFAR10, CIFAR100~\cite{krizhevsky2009learning}, CINIC10~\cite{darlow2018cinic}, and TinyImageNet~\cite{le2015tiny}.

\vspace{2mm}\noindent{\bf Models.}
We used ResNet18~\cite{he2015deep}, VGG16~\cite{simonyan2015deep}, EfficientNetV2-S~\cite{tan2021efficientnetv2}, and small MobileNetV3~\cite{howard2019searching}.

\vspace{2mm}\noindent{\bf Backdoors.}
We considered eight different triggers applied with three state-of-the-art attack methods: six different patch triggers with BadNets~\cite{gu2019badnets}, an invisible trigger with Steganography~\cite{li2020invisible}, and an optimized trigger with Narcissus~\cite{zeng2023narcissus}. Among the methods, Narcissus is the only one that solely infects target class samples, thus making it more stealthy without requiring a label change. Examples of the applied triggers are visualized in Figure~\ref{fig:trigger}.

\vspace{2mm}\noindent{\bf Baselines.}
We considered four suitable backdoor unlearning methods for comparison, two of which require \ac{bdk}: (1) fine-tuning, which penalizes a high difference between activations of clean and poisoned data (actFT)~\cite{qiao2019defending}, and (2) BaEraser, which uses gradient ascent for unlearning ~\cite{liu2022backdoor}. The other two methods work with $\neg$\ac{bdk}: (3) fine-tuning on clean $D_U$ in the same way as the initial training (basicFT), and (4) \ac{nad}, a knowledge distillation approach where the basicFT model, acting as the teacher model, only passes on its ability to clean data~\cite{li2021neural}.

\vspace{2mm}\noindent{\bf Evaluation Metrics.}
For evaluation, we used the \ac{asr}, which is the ratio of a backdoor sample being misclassified as $y_t$, Clean Test Accuracy (ACC), and \ac{ctca}, which is the ACC for samples of class $y_t$. We were interested in examining the change in \ac{ctca} because our unlearning method often leads to a drop in \ac{ctca} alongside \ac{asr}. Repairing can mitigate this side effect.

We introduce a two-part scoring function to estimate the forgetting and utility quality after unlearning combined in one value. 
The forgetting quality is estimated by subtracting the \ac{asr} ratio of the unlearned (U) and victim model (V) from 1. A higher drop in \ac{asr} after unlearning indicates a higher score for the forgetting part. The ACC ratio of the unlearned and the retrained model (R), which is trained on $D_C$ from scratch, estimates the utility part. We strive to achieve the same or even higher ACC on the unlearned model compared to the retrained model that was never poisoned before. We use the retrained model for this ratio because, especially in cases with a high poisoning rate $\rho$, the poisoning can negatively influence the ACC of the victim model, thus not representing a clean model performance. The final score value is calculated as

\begin{equation}\label{eq:score}
\textrm{Score} = (1-\frac{\textrm{ASR}^U}{\textrm{ASR}^V}) \cdot \frac{\textrm{ACC}^U}{\textrm{ACC}^R}.
\end{equation}

\vspace{2mm}\noindent{\bf Base Configuration.}
We used specific base configurations if not stated otherwise for an experiment. Experiments were performed on the CIFAR10 dataset and ResNet18 as the victim model. The training dataset $D_T$ was infected with a poisoning rate $\rho$ of 5\%, with the trigger displayed in Figure~\ref{fig:white}. The backdoor target class $y_t$ was two, representing birds. 
Unlearn dataset $D_U$ consisted of 5000 samples, but our method only used 512 by default.

\subsection{Results}
We examined the performance of our unlearning in different settings.

\vspace{2mm}\noindent{\bf Different Datasets.}
In this experiment, we trained ResNet18 models poisoned with backdoor triggers on five datasets: MNIST, CIFAR10, CIFAR100, CINIC10, and TinyImageNet. Table~\ref{tab:datasets} summarizes the evaluation of the proposed unlearning method with the different datasets in terms of \ac{asr}, ACC, and \ac{ctca}. The proposed method effectively reduced the \ac{asr} on every dataset, except MNIST (grayscale images), when we had $\neg$\ac{bdk}. Repairing improved ACC for several datasets and restored \ac{ctca} while increasing \ac{asr} by a lesser extent. There was an exclusively negative influence on performance with CINIC10 repairing.

\vspace{2mm}\noindent{\bf Different Models.}
In this experiment, we trained different models: ResNet18, VGG16, EfficientNetV2-S, and MobileNetV3 (small version). Table~\ref{tab:models} presents the performance of the proposed unlearning method with the different models. The proposed method with \ac{bdk} was effective on every tested model. After unlearning, we retained a good ACC on EfficientNetv2 with \ac{bdk}, while the utility was lost with $\neg$\ac{bdk}. However, repairing both models resulted in similar final performance, which benefited $\neg$\ac{bdk} but decreased performance for \ac{bdk}.

\vspace{2mm}\noindent{\bf Comparison with State-of-the-Art Methods.}
In this experiment, we trained models and performed unlearning with different backdoor triggers. As described in Section~\ref{sec:setup}, we considered four baseline methods: actFT and BaEraser under \ac{bdk}, and basicFT and \ac{nad} under $\neg$\ac{bdk} with eight poison triggers for comparison. The models were trained with different poisoning budgets $\rho$ and $alpha$ values of the RGBA-coded trigger. Figure~\ref{fig:trigger} depicts the triggers.

Tables~\ref{tab:trigger1} and \ref{tab:trigger2} summarize the performance of the proposed unlearning method with the different baseline methods in terms of score (see Section~\ref{sec:setup}). The score metric measured the forgetting and utility quality after unlearning. For backdoor verification, the $alpha$ value was multiplied by three up to a maximum of 100\%. For actFT to be effective, we multiplied the $alpha$ value for unlearning by the same magnitude. Our method outperformed the previous methods in terms of score with or without \ac{bdk} for most triggers.

\subsection{Analysis} 

We analyze the proposed unlearning method in terms of sample efficiency, time efficiency, and potential backdoor detection application.

\vspace{2mm}\noindent{\bf Sample Efficiency.}
Table~\ref{tab:samplecount} shows the performance of the proposed unlearning method when using different numbers of samples for unlearning. With \ac{bdk}, the performance did not depend on the sample count. With $\neg$\ac{bdk}, the performance with different sample counts did not follow a clear pattern. Notably, the \ac{asr} with two samples was exceptionally high compared with others. Therefore, we recommend using a minimum of four samples for unlearning with $\neg$\ac{bdk}.

Table~\ref{tab:time} presents a performance comparison of the proposed unlearning and state-of-the-art methods in terms of several metrics, including the time required for unlearning. The baseline methods compared with ours depended more on a high sample count in $D_U$. 
For most of the baselines, more samples resulted in a higher score. An exception is BaEraser, which had the best performance with 500 samples.

\vspace{2mm}\noindent{\bf Time Efficiency.}
In the particular scenario where training data is available and retraining is feasible, assessing the computational cost saved with unlearning compared with retraining is an important metric. When unlearning is not drastically more time efficient, retraining is the preferred choice to perfectly remove the influence of the data we want to forget. The unlearning time in our scenario with training data unavailability is not a deciding factor. Still, we have to consider the trade-off between unlearning performance and the cost of computing for the benefit of scalability.

As evident in Table~\ref{tab:time}, our method requires significantly less time and fewer samples for unlearning than other methods. Our method uses only a single forward pass to extract the activation, and the remaining operations are simple matrix operations. In comparison, all baseline methods require optimization with backpropagation, which generally is more computationally expensive, resulting in a higher unlearning time. 

\vspace{2mm}\noindent{\bf Target Class Detection.}
Our experiments show that the proposed unlearning method reduced the backdoor class accuracy (\ac{ctca}). To address this issue, we introduce a repair step to preserve utility. Before repairing, we can utilize significant decreases in target class accuracy with $\neg$\ac{bdk} to detect a backdoor and the target class. We carried out a simple experiment on poisoned models on all ten classes of CIFAR10. We can usually observe an unusual decrease in accuracy for a single class. When we assumed the single class as the target class, we got a target class prediction accuracy of $80\%$. A formula sets the accuracy of the different classes into relation and returned a classification value. Comparing the value to a threshold value gives us a binary prediction for the existence of a backdoor. The backdoor detection accuracy was $67\%$ when poisoned and $80\%$ when having a clean model.

\section{Discussion}
We demonstrated a model-editing method that unlearns the backdoor trigger feature embedded in a backdoored model by utilizing the activation of clean or poisoned samples. Our method achieves consistent unlearning performance across various settings with different models, datasets, and backdoor triggers by state-of-the-art attacks. Apart from the unlearning performance, there are two key factors where our method exceeds current state-of-the-art methods by a significant margin. Our unlearning process is exceptionally fast to compute and, most of the time, requires only a handful of samples to unlearn the backdoor effectively. Additionally, we can use information gained after unlearning for backdoor presence and target class prediction.

We experimented with our algorithm and found specific activation-manipulating formulas that gave us the best unlearning performance for model editing. In Equation~\ref{eq:neg}, negating poisoned activation with \ac{bdk} and clean activation with $\neg$\ac{bdk} worked the best. With \ac{bdk}, we negate the activation of the trigger-infected data we want to forget. Previously, Ilharco et al.~\cite{ilharco2023editing} arrived at the same conclusion as we did, that moving in the negative direction of extracted information can lead to unlearning.

The most significant limitation of our method is that it disturbs the overall utility and unlearns the targeted class instead of only backdoor samples. Therefore, repairing is used to restore lost utility.
The experimental scope was limited, and we covered only convolutional neural networks. 

Hence, for future work, we shall explore the unlearning method with different architectures, such as vision transformers~\cite{dosovitskiy2021image}, mixers~\cite{tolstikhin2021mlpmixer}, \etc We shall investigate explainability methods to better understand the parts of the algorithm that are responsible for effective unlearning and ideally improve the unlearning performance without loss of utility. In addition, not limiting the method to backdoor unlearning, we shall expand the applications of unlearning, such as privacy-related unlearning applications. In this work, we analyzed backdoors in images, but for future work, we shall expand experiments to other data types, such as text or audio data.


\section{Conclusion}
Our method offers a new approach to tackling the security issue posed by backdoor attacks by mitigating the influence of the attacks on a backdoor-infected model without requiring access to the original training data. Multiple experiments show the broad applicability of our method in various settings. It performs better than previous backdoor unlearning methods in most scenarios. Moreover, it executes faster and requires fewer samples for unlearning than the previous methods.

\section*{Acknowledgements}
This work was partially supported by JSPS KAKENHI Grants JP21H04907 and JP24H00732, and by JST CREST Grants JPMJCR18A6 and JPMJCR20D3 including AIP challenge program, and by JST AIP Acceleration Grant JPMJCR24U3 Japan.

%
%
\bibliographystyle{splncs04}
\bibliography{main}


\appendix

\setcounter{figure}{3}
\setcounter{table}{6}
\setcounter{equation}{6}

\section{Experimental Setup}

\subsection{System Hardware}

We used three different setups for our experiments due to changing resource accessibility. First one with  251 GiB system memory, four Intel Xeon E5-2698 v4 CPUs~\cite{CPU}, and one NVIDIA Tesla V100~\cite{GPU}. The second one with 264 GiB system memory, 36 Intel Core i9-10980XE CPUs~\cite{CPU2}, and one NVIDIA RTX A6000~\cite{GPU2}. The third one contained 16 GiB system memory, one AMD Ryzen 5 5600X CPU~\cite{CPU3}, and one Nvidia RTX 4600 Ti GPU~\cite{GPU3}.

\begin{table}[htbp]
\caption{Efficiency of proposed unlearning compared with state-of-the-art methods. We experimented with 50\%(5000), 5\%(500), 0.5\%(50), and 0.05\%(5) of unseen CIFAR10 data for unlearning. Best results are highlighted in bold.}
\centering
\ra{1.1}
\resizebox{\linewidth}{!}{%
\begin{tabular}{lcccccc}\toprule
\multirow{2}{*}{\textbf{Method}} &Number of& \multicolumn{5}{c}{Metric} \\
\cmidrule(lr){3-7}
&Samples& Score ($\uparrow$) & ASR ($\downarrow$) & ACC ($\uparrow$)& CTCA ($\uparrow$) & Time ($\downarrow$)\\
\midrule
\multirow{4}{*}{actFT~\cite{qiao2019defending}}&5000& \textbf{92.95$\pm$4.72} & 5.28$\pm$2.48 & \textbf{69.64$\pm$0.92} & 58.1$\pm$9.1 & 3.96$\pm$0.34\\
&500& 7.0$\pm$2.09 & 87.17$\pm$2.86 & \textbf{69.41$\pm$1.02} & \textbf{61.47$\pm$7.99} & 2.81$\pm$0.04\\
&50& 0.21$\pm$0.06 & 93.67$\pm$1.59 & \textbf{69.81$\pm$1.04} & \textbf{61.42$\pm$8.41} & 3.01$\pm$0.5\\
&5& 0.22$\pm$0.08 & 93.66$\pm$1.6 & \textbf{69.82$\pm$1.06} & \textbf{61.58$\pm$8.38} & 2.44$\pm$0.03\\
\cmidrule(lr){2-7}
\multirow{4}{*}{BaEraser~\cite{liu2022backdoor}}&5000& 57.47$\pm$15.65 & 2.0$\pm$2.79 & 41.81$\pm$12.3 & 14.79$\pm$20.77 & 623.3$\pm$111.46\\
& 500 & 80.69$\pm$1.93 & 3.81$\pm$2.41 & 59.45$\pm$0.83 & 32.98$\pm$1.1 & 138.88$\pm$2.87\\
&50& 54.48$\pm$13.21 & 24.16$\pm$16.46 & 51.83$\pm$2.66 & 39.64$\pm$8.6 & 21.35$\pm$0.17\\
&5& 16.88$\pm$11.53 & 43.01$\pm$34.71 & 19.06$\pm$4.29 & 56.76$\pm$31.71 & 20.23$\pm$0.2\\
\cmidrule(lr){2-7}
\multirow{4}{*}{Ours(BDK)}&5000& \textbf{92.38$\pm$1.67} & \textbf{0.0$\pm$0.0} & 65.39$\pm$0.42 & 0.0$\pm$0.0 & \textbf{0.81$\pm$0.05} \\
&500& \textbf{92.3$\pm$1.45} & \textbf{0.0$\pm$0.0} & 65.2$\pm$0.58 & 0.0$\pm$0.0 & \textbf{0.58$\pm$0.01} \\
& 50 & \textbf{92.4$\pm$1.39} & \textbf{0.0$\pm$0.0} & 65.29$\pm$0.58 & 0.0$\pm$0.0 & \textbf{0.38$\pm$0.01} \\
&5& \textbf{92.51$\pm$1.34} & \textbf{0.0$\pm$0.0} & 65.37$\pm$0.56 & 0.0$\pm$0.0 & \textbf{0.38$\pm$0.02} \\
\cmidrule(lr){2-7}
&5000& 77.95$\pm$1.6 & 9.1$\pm$1.08 & 61.11$\pm$0.89 & 43.98$\pm$3.66 & 49.66$\pm$0.53\\
Ours(BDK)&500& 76.63$\pm$3.63 & \textbf{0.01$\pm$0.0} & 54.14$\pm$2.4 & 0.0$\pm$0.0 & 67.7$\pm$1.76\\
+repair& 50 & 83.64$\pm$2.81 & \textbf{0.0$\pm$0.0} & 59.09$\pm$1.82 & 0.0$\pm$0.0 & 72.02$\pm$0.98 \\
&5& 67.56$\pm$8.98 & \textbf{0.0$\pm$0.0} & 47.69$\pm$5.95 & 0.0$\pm$0.0 & 70.39$\pm$1.25 \\
\cmidrule(lr){1-7}
\multirow{4}{*}{basicFT}&5000& 69.49$\pm$1.11 & 6.0$\pm$1.07 & 52.57$\pm$0.91 & 33.86$\pm$4.07 & 71.56$\pm$1.47\\
&500& 14.14$\pm$0.1 & \textbf{0.0$\pm$0.0} & 9.99$\pm$0.03 & 0.0$\pm$0.0 & 73.34$\pm$0.2\\
&50& 14.13$\pm$0.1 & \textbf{0.0$\pm$0.0} & 9.98$\pm$0.02 & 0.0$\pm$0.0 & 71.59$\pm$0.25\\
&5& 14.17$\pm$0.11 & \textbf{0.0$\pm$0.0} & 10.01$\pm$0.0 & 0.0$\pm$0.0 & 71.49$\pm$0.03\\
\cmidrule(lr){2-7}
\multirow{4}{*}{NAD~\cite{li2021neural}}& 5000& 71.23$\pm$3.2 & 4.54$\pm$1.48 & 52.95$\pm$1.41 & 32.05$\pm$1.3 & 114.9$\pm$1.45\\
&500& 42.18$\pm$5.42 & 6.68$\pm$2.06 & 32.09$\pm$3.97 & 23.18$\pm$9.4 & 117.69$\pm$1.9\\
&50& 42.49$\pm$7.39 & 9.36$\pm$5.28 & 33.58$\pm$6.54 & 20.92$\pm$11.71 & 111.91$\pm$1.06\\
&5& 7.44$\pm$8.48 & 59.03$\pm$41.63 & 11.71$\pm$1.64 & \textbf{63.93$\pm$37.39} & 118.7$\pm$5.99\\
\cmidrule(lr){2-7}
\multirow{4}{*}{Ours($\neg$BDK)}&5000& 82.67$\pm$0.51 & \textbf{0.0$\pm$0.0} & 58.53$\pm$0.97 & 0.0$\pm$0.0 & \textbf{0.81$\pm$0.02} \\
&500& 82.8$\pm$0.53 & \textbf{0.0$\pm$0.0} & 58.49$\pm$0.21 & 0.0$\pm$0.0 & \textbf{0.51$\pm$0.06} \\
& 50 & \textbf{89.9$\pm$2.02} & \textbf{0.0$\pm$0.0} & \textbf{63.52$\pm$1.24} & 0.0$\pm$0.0 & \textbf{0.36$\pm$0.02} \\
&5& \textbf{89.5$\pm$1.72} & \textbf{0.0$\pm$0.0} & \textbf{63.23$\pm$0.86} & 0.0$\pm$0.0 & \textbf{0.4$\pm$0.01} \\
\cmidrule(lr){2-7}
&5000& 77.47$\pm$2.76 & 10.79$\pm$4.74 & 62.09$\pm$0.81 & 51.07$\pm$4.78 & 49.4$\pm$0.17\\
Ours($\neg$BDK)&500& 56.7$\pm$14.54 & 24.66$\pm$18.58 & 54.72$\pm$2.02 & 45.26$\pm$9.69 & 68.96$\pm$0.33\\
+repair& 50 & 83.53$\pm$1.0 & \textbf{0.0$\pm$0.0} & 59.02$\pm$0.64 & 0.0$\pm$0.0 & 73.86$\pm$1.75 \\
&5& 62.18$\pm$4.95 & \textbf{0.04$\pm$0.04} & 43.93$\pm$3.13 & 0.07$\pm$0.09 & 74.43$\pm$0.43 \\
\cmidrule(lr){1-7}
Retraining&47500 & -& 4.03$\pm$0.99 & 70.27$\pm$ 0.66 & 	60.55$\pm$4.3 & 423.56$\pm$73.96\\
\bottomrule
\end{tabular}
}
\label{tab:time2}
\end{table}

\section{Additional Results}

\subsection{Sample and Time Efficency}

This section expands upon the limited results shown in Table~\ref{tab:time}. Table~\ref{tab:time2} shows the full results of the experiment and, specifically, the performance of our method when we use repairing. With \ac{bdk}, repairing required 5000 samples when aiming to increase \ac{ctca} while maintaining a relatively low \ac{asr}. With $\neg$\ac{bdk}, repairing had a positive effect on \ac{ctca} with 500 samples, but with 5000, the trade-off between unlearning and target class utility is better. 

The baseline methods depended heavily on a certain number of samples to achieve method-specific top performance. Certain baseline methods, such as actFT and basicFT, seem to require a high number of samples to have any effect in reducing ASR or maintaining ACC above random guessing.
Without repairing, our method performed similarly for different test runs with different-sized unlearning data. 

Notably, when comparing the results from actFT and basicFT with five samples, we can see a problem with our current scoring function. While actFT, with a score close to 0, has not unlearned the backdoor but maintains model utility, basicFT achieves a higher score with a model that has completely lost its utility. Our current scoring function prioritizes successful backdoor mitigation over maintaining model utility.
Additionally, the scoring function does not emphasize tracking the complete loss of utility for a single class. Maintaining at least a decent ACC for each class is crucial to ensuring that a model retains its essential utility.

\subsection{Mitigating Limitation with Repairing}

This experiment shows how the learning rate influences repairing with one epoch. We aim for a suitable learning rate to restore utility without sacrificing too much backdoor unlearning capability.
We are limited to finetune only on unseen unlearn data. This limitation can lead to reduced generalizability compared to the original model because $D_U$ is one order of magnitude smaller than $D_T$.

\begin{table}[ht]
\caption{Repairing performance with different values for learning rate $\eta$. For comparison, we include the model state before unlearning (Before) and after weight editing without repairing (None). Best results are highlighted in bold.}
\centering
\ra{1.1}
\begin{tabular}{c@{\hskip 0.2in}c@{\hskip 0.2in}c@{\hskip 0.2in}c@{\hskip 0.2in}c}
\toprule
& $\eta$ & ASR ($\downarrow$) & ACC ($\uparrow$) & CTCA ($\uparrow$)\\
\midrule
Before & - & 94.53$\pm$1.29 & 70.5$\pm$0.18 & 61.39$\pm$8.51\\
\midrule
\multirow{6}{*}{BDK} & None & \textbf{0.0$\pm$0.0} & 65.49$\pm$0.39 & 0.0$\pm$0.0\\
& 1e-6 & \textbf{0.0$\pm$0.0} & 62.29$\pm$0.09 & 0.0$\pm$0.0\\
&1e-5 & \textbf{0.0$\pm$0.0} & 62.6$\pm$0.1 & 0.0$\pm$0.0\\
&1e-4 & \textbf{0.0$\pm$0.0} & 64.08$\pm$0.31 & 0.0$\pm$0.0\\
&1e-3 & 17.83$\pm$5.28 & \textbf{67.56$\pm$0.23} & \textbf{47.23$\pm$2.78}\\
&1e-2 & \textbf{9.79$\pm$0.88} & 59.87$\pm$3.04 & \textbf{42.07$\pm$7.72}\\
&1e-1 & \textbf{0.3$\pm$0.43} & 20.44$\pm$1.56 & 0.87$\pm$1.23\\
\midrule
\multirow{6}{*}{$\neg$BDK}& None& \textbf{0.3$\pm$0.21}  & 59.18$\pm$2.63 & 0.14$\pm$0.19\\
& 1e-6 & 28.77$\pm$23.08 & \textbf{62.59$\pm$0.41} & 3.3$\pm$3.85\\
&1e-5 & 32.03$\pm$24.75 & \textbf{63.18$\pm$0.56} & 5.77$\pm$5.66\\
&1e-4 & 43.17$\pm$30.91 & \textbf{66.08$\pm$1.47} & 20.56$\pm$14.76\\
&1e-3 & 53.48$\pm$12.29 & \textbf{68.47$\pm$0.17} & \textbf{55.06$\pm$2.67}\\
&1e-2 & \textbf{23.09$\pm$8.07} & \textbf{61.83$\pm$1.33} & \textbf{55.02$\pm$4.41}\\
&1e-1 & \textbf{6.01$\pm$3.54} & 24.57$\pm$0.38 & 13.66$\pm$8.73\\
\bottomrule
\end{tabular}
\label{tab:repair}
\end{table}

Table~\ref{tab:repair} shows repairing can help restore \ac{ctca} while risking increasing the \ac{asr} and potentially decreasing the ACC for other classes. The effectiveness of repairing depends heavily on the used dataset and poisoned trigger. In this experiment, the learning rate 1e-2 was the most effective when aiming for a decent \ac{ctca} and a low \ac{asr}, achieving a good balance between forgetting and utility.

\subsection{Effect of Activation Mean and Standard Deviation}
\label{sec:meanstd}

This experiment shows the best values for mean and standard deviation hyperparameters $\lambda$ and $\gamma$ for achieving effective unlearning. We also show how changing the \ac{bn} \ac{ma} parameters during activation extraction affects the unlearning performance. 

\begin{figure}[ht]
\centering
\subfloat[]{\includegraphics[width=0.49\linewidth]{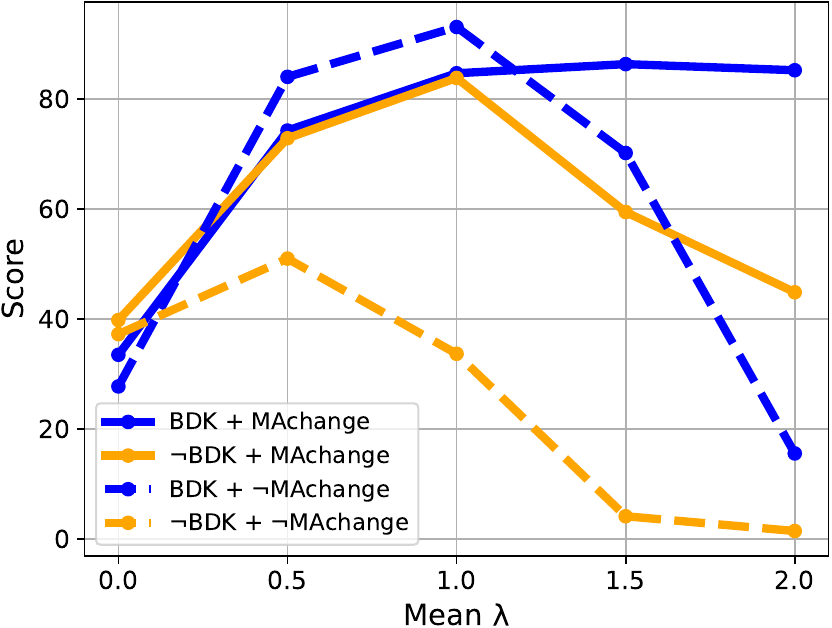}\label{fig:meanAblation}}
\hfil
\subfloat[]{\includegraphics[width=0.49\linewidth]{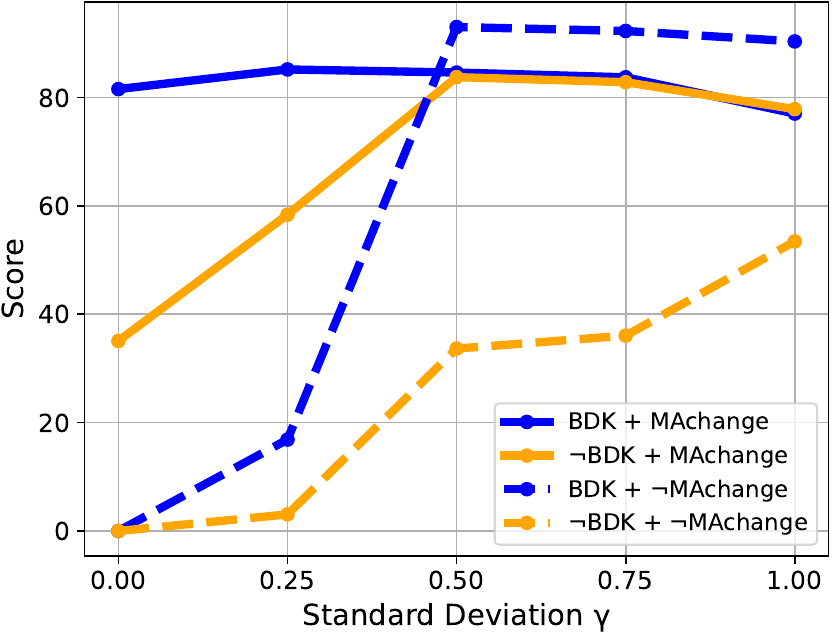}\label{fig:stdAblation}}
\caption{Performance of our method when using different values for hyperparameters $\lambda$ and $\gamma$. Compared with or without change in \ac{bn} \ac{ma} parameters.}
\label{fig:meanstd}
\end{figure}

Figure~\ref{fig:meanstd} shows that allowing change of \acp{ma} can improve performance when we unlearn with $\neg$\ac{bdk}. For unlearning with \ac{bdk}, we can achieve the best performance when we keep the \acp{ma} fixed. When using the respective better-performing process, we get the best performance with 1.0 as a value for $\lambda$. 

Parameter $\gamma$ mainly influences the impact of the unlearning process on the model weights. A higher value for $\gamma$ generally means a more invasive model editing and a substantial weight change of those neurons that reacted strongly to the input dataset. 

With \ac{bdk}, we achieve the best performance with fixed \acp{ma}. Without \ac{bdk}, allowing change of \acp{ma} is better. 
Rescaling with a mean of 1.0 and standard deviation of 0.5 is most suitable for stable, high results.
When $\gamma$ equals 0, we can observe the isolated influence of the change in \ac{ma} parameters. The reason is that the influence of model editing is nonexistent when activations are scaled to 0 before shifting every activation value to 1.

\subsection{Effects of different Activations}
\label{sec:actform}

In this experiment, we show how the unlearning performs when we choose a different activation formula for Equation~\ref{eq:neg}. Instead of using just the negative activation of the unlearn dataset $D_U$, we introduce two hyperparameters $\alpha$ and $\beta$ that balance out the activation with clean and poisoned $D_U$. In the case of $\neg$\ac{bdk}, where we have no trigger for poisoning, clean and poisoned activation are the same. The formula that replaces Equation~\ref{eq:neg} is calculated as

\begin{equation}\label{eq:act}
A = \alpha \cdot A_\text{clean} + \beta \cdot A_\text{poisoned}.
\end{equation}

\begin{table}[ht]
\caption{Score ($\uparrow$) comparison for different values for hyperparameter $\alpha$ and $\beta$. Best results are highlighted in bold.}
\centering
\ra{1.3}
\begin{tabular}{c@{\hskip 0.2in}c@{\hskip 0.2in}c@{\hskip 0.2in}c@{\hskip 0.2in}c}\toprule
        &\multirow{2}{*}{$\beta$} & \multicolumn{3}{c}{$\alpha$}\\
        \cmidrule(lr){3-5}
        && -1 & 0 & 1\\
        \midrule
        \multirow{3}{*}{BDK} & -1 & \textbf{93.38$\pm$0.58} & \textbf{93.02$\pm$0.66} & \textbf{92.89$\pm$1.17}\\
        & 0& 33.67$\pm$42.76 & 14.34$\pm$0.16 & 0.0$\pm$0.0\\
        & 1& 0.0$\pm$0.0 & 0.0$\pm$0.0 & 0.0$\pm$0.0\\
        \midrule
        \multirow{3}{*}{$\neg$BDK} & -1 & \textbf{86.67$\pm$1.43} & \textbf{86.67$\pm$1.43} & 14.34$\pm$0.16\\
        & 0& \textbf{86.67$\pm$1.43} & 14.34$\pm$0.16 & 0.62$\pm$0.88\\
        & 1& 14.34$\pm$0.16 & 0.62$\pm$0.88 & 0.62$\pm$0.88\\
        \bottomrule
    \end{tabular}
\label{tab:formula}
\end{table}

Table~\ref{tab:formula} shows that our method performs well in both unlearning scenarios when we set $\beta$ to -1 and $\alpha$ to either -1 or 0. For the final algorithm, we decided to use values 0 and -1 for $\alpha$ and $\beta$, respectively, because setting at least one hyperparameter to 0 reduces the complexity of the function and allows us to extract one activation set less.

\subsection{Target Class Dependencies}

In this experiment, we compare the performance of our methods with the baseline methods when we use different backdoor target classes. The robustness of the classes can vary significantly. We want to show that our method achieves consistency for all classes in CIFAR10.

\begin{table}[ht]
\caption{Score ($\uparrow$) comparison of different methods based on the backdoor target class. Best results are highlighted in bold.}
\centering
\ra{1.1}
\resizebox{\linewidth}{!}{%
\begin{tabular}{c@{\hskip 0.1in}c@{\hskip 0.1in}c@{\hskip 0.1in}c@{\hskip 0.1in}c@{\hskip 0.1in}c@{\hskip 0.1in}c}\toprule
    \textbf{Target}& \multicolumn{3}{c}{BDK} & \multicolumn{3}{c}{$\neg$BDK}\\
    \cmidrule(lr){2-4}\cmidrule(lr){5-7}  
    \textbf{Class} & actFT~\cite{qiao2019defending} & BaEraser~\cite{liu2022backdoor} & Ours& basicFT & NAD~\cite{li2021neural} & Ours\\
    \midrule
    0 & \textbf{94.85$\pm$0.57} & 49.24$\pm$10.22 & 90.1$\pm$0.64 & 70.66$\pm$5.0 & 68.71$\pm$3.77 & \textbf{80.82$\pm$2.44}\\
    1 & \textbf{94.49$\pm$1.49} & 43.47$\pm$21.12 & 88.28$\pm$0.35 & 58.32$\pm$6.92 & 68.26$\pm$0.72 & \textbf{75.57$\pm$0.62}\\
    2 & \textbf{94.62$\pm$2.61} & 59.5$\pm$4.62 & 93.02$\pm$0.66 & 61.9$\pm$3.63 & 65.33$\pm$3.04 & \textbf{83.77$\pm$3.34}\\
    3 & \textbf{91.21$\pm$1.83} & 31.77$\pm$10.03 & 94.41$\pm$0.17 & 49.96$\pm$11.41 & 65.98$\pm$5.62 & \textbf{87.46$\pm$0.11}\\
    4 & \textbf{93.71$\pm$3.22} & 57.17$\pm$25.75 & 90.61$\pm$2.5 & 62.76$\pm$5.06 & 68.32$\pm$1.17 & \textbf{87.63$\pm$0.5}\\
    5 & \textbf{96.15$\pm$3.24} & 61.39$\pm$15.54 & 93.78$\pm$1.7 & 57.12$\pm$18.84 & 69.7$\pm$4.58 & \textbf{89.67$\pm$0.77}\\
    6 & \textbf{93.05$\pm$2.45} & 38.2$\pm$20.33 & 89.08$\pm$0.83 & 55.08$\pm$18.59 & 67.38$\pm$3.96 & \textbf{81.53$\pm$2.21}\\
    7 & \textbf{95.86$\pm$2.12} & 29.57$\pm$14.03 & 90.78$\pm$1.76 & 70.93$\pm$3.53 & 71.59$\pm$4.47 & \textbf{77.7$\pm$3.05}\\
    8 & \textbf{93.37$\pm$2.73} & 35.28$\pm$5.39 & 87.23$\pm$1.2 & 30.56$\pm$23.24 & 66.2$\pm$2.52 & \textbf{77.32$\pm$0.47}\\
    9 & \textbf{92.08$\pm$3.09} & 53.83$\pm$13.81 & 90.57$\pm$0.91 & 58.7$\pm$13.01 & 70.75$\pm$5.21 & \textbf{85.46$\pm$1.6}\\
    \midrule
    Average & \textbf{93.94$\pm$2.33} & 45.94$\pm$14.08 & 90.79$\pm$1.07 & 57.6$\pm$10.92 & 68.22$\pm$3.51 & \textbf{82.69$\pm$1.51}\\
    \bottomrule
    \end{tabular}
    }
\label{tab:targetclass}
\end{table}

Table~\ref{tab:targetclass} shows that our method performs better on average when we have $\neg$\ac{bdk}. With \ac{bdk}, the actFT proposed by \cite{qiao2019defending} performs slightly better on average.


\begin{acronym}

\acro{asr}[ASR]{Attack Success Rate}
\acro{bdk}[BDK]{Backdoor Knowledge}
\acro{bn}[BN]{Batch Normalization}
\acro{ctca}[CTCA]{Clean Target Class Accuracy}
\acro{ma}[MA]{Moving Average}
\acro{nad}[NAD]{Neural Attention Distillation}

\end{acronym}
\end{document}